\documentclass[letterpaper,10pt,conference]{ieeeconf}
\IEEEoverridecommandlockouts
\usepackage{graphicx}
\usepackage{amsmath,amssymb}
\usepackage{booktabs}
\usepackage{array}
\usepackage{multirow}
\usepackage{xcolor}
\usepackage{url}
\usepackage{eso-pic}
\newcommand{\mechanicalrepo}{\url{https://github.com/mskim1309/ICRA-2027-Master-Console}}

\usepackage{tikz}
\usepackage{caption}
\usepackage{cuted}
\usepackage{placeins}
\definecolor{pframe}{gray}{0.72}

\newlength{\ph}

\newcommand{\panelh}[3]{%
  \begin{tikzpicture}[inner sep=0pt]
    \node (I) {\includegraphics[height=#1]{#2}};
    \draw[pframe,line width=0.4pt] (I.south west) rectangle (I.north east);
    \node[anchor=north west, fill=black, fill opacity=0.75, text opacity=1,
          text=white, font=\bfseries\small, inner xsep=4.5pt, inner ysep=3.5pt]
          at ([xshift=1.6pt,yshift=-1.6pt]I.north west) {#3};
  \end{tikzpicture}}

\newcommand{\paneltrimh}[4]{%
  \begin{tikzpicture}[inner sep=0pt]
    \node (I) {\includegraphics[height=#1,trim=#4,clip]{#2}};
    \draw[pframe,line width=0.4pt] (I.south west) rectangle (I.north east);
    \node[anchor=north west, fill=black, fill opacity=0.75, text opacity=1,
          text=white, font=\bfseries\small, inner xsep=4.5pt, inner ysep=3.5pt]
          at ([xshift=1.6pt,yshift=-1.6pt]I.north west) {#3};
  \end{tikzpicture}}

\title{\LARGE \bf
Surgical Master Console Using General-Purpose Robot Arms and a Separable Articulated Distal Interface: Porcine In-Vivo Evaluation
}

\author{%
\small
Minsung Kim$^{*,1,2,4}$,
Seonho Shim$^{*,2,9}$,
Dongho Yee$^{1,2,4,5}$,
Younghoon Noh$^{2,4}$,\\
Juahn Oh$^{1,2,8}$,
Yechan Seo$^{1,2,3}$,
Jinseok Lee$^{2,4}$,
Jiyul Lee$^{1,2,3}$,\\
Seong Jeong$^{1,2,3}$,
Hyuk Choi$^{1,2,3}$,
Youngbin Kong$^{1,7}$,
and Hyoun-Joong Kong$^{\dagger,1,3,6}$%
\thanks{\scriptsize *Co-first authors. \textdagger Corresponding author.}%
\thanks{\scriptsize $^{1}$Department of Transdisciplinary Medicine, Seoul National University Hospital, Seoul, Republic of Korea.
$^{2}$Rosota Inc., Seoul, Republic of Korea.
$^{3}$Department of Medicine, Seoul National University College of Medicine, Seoul, Republic of Korea.
$^{4}$Department of Mechanical Engineering, Seoul National University, Seoul, Republic of Korea.
$^{5}$Department of Computer Science and Engineering, Seoul National University, Seoul, Republic of Korea.
$^{6}$Institute of Convergence Medicine with Innovative Technology, Seoul National University Hospital, Seoul, Republic of Korea.
$^{7}$Interdisciplinary Program in Medical Informatics, Seoul National University College of Medicine.
$^{8}$Eulji University College of Medicine, Daejeon, Republic of Korea.
$^{9}$Department of Mechanical Engineering, Chungang University, Seoul, Republic of Korea.}%
}

\newcommand{\ieeesubmissionnotice}{%
  \AddToShipoutPictureFG*{%
    \AtPageLowerLeft{%
      \hspace{0.5\paperwidth}%
      \raisebox{0.45in}{%
        \makebox(0,0){\footnotesize This work has been submitted to the IEEE for possible publication. Copyright may be transferred without notice, after which this version may no longer be accessible.}%
      }%
    }%
  }%
}

\begin{document}
\maketitle
\ieeesubmissionnotice
\thispagestyle{empty}
\pagestyle{empty}

\begin{strip}
  \vspace*{-32mm}
  \centering
  \setlength{\tabcolsep}{1.5pt}
  \setlength{\ph}{34mm}
  \resizebox{\textwidth}{!}{%
  \begin{tabular}{@{}cccc@{}}
    \panelh{\ph}{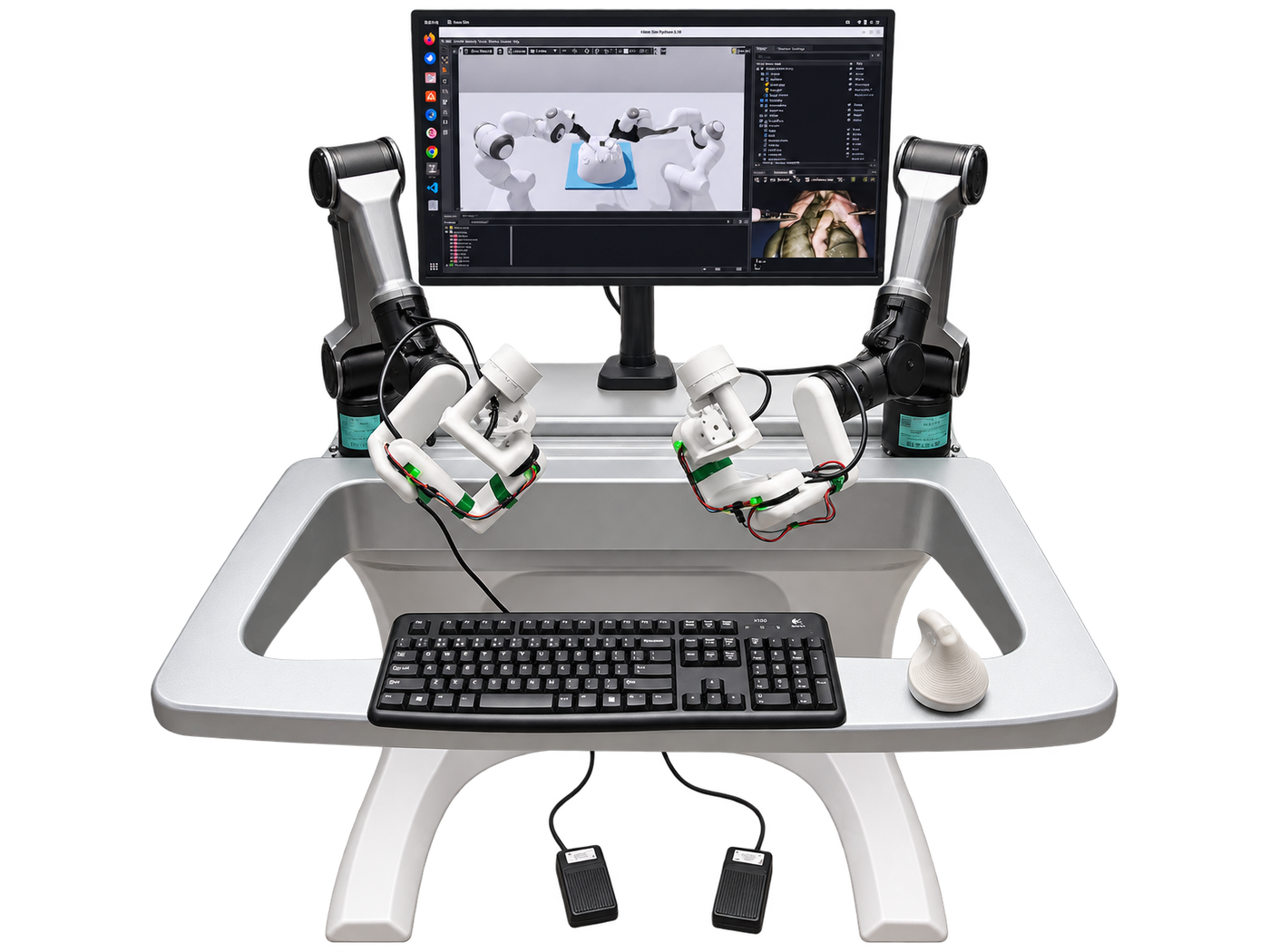}{(a)} &
    \panelh{\ph}{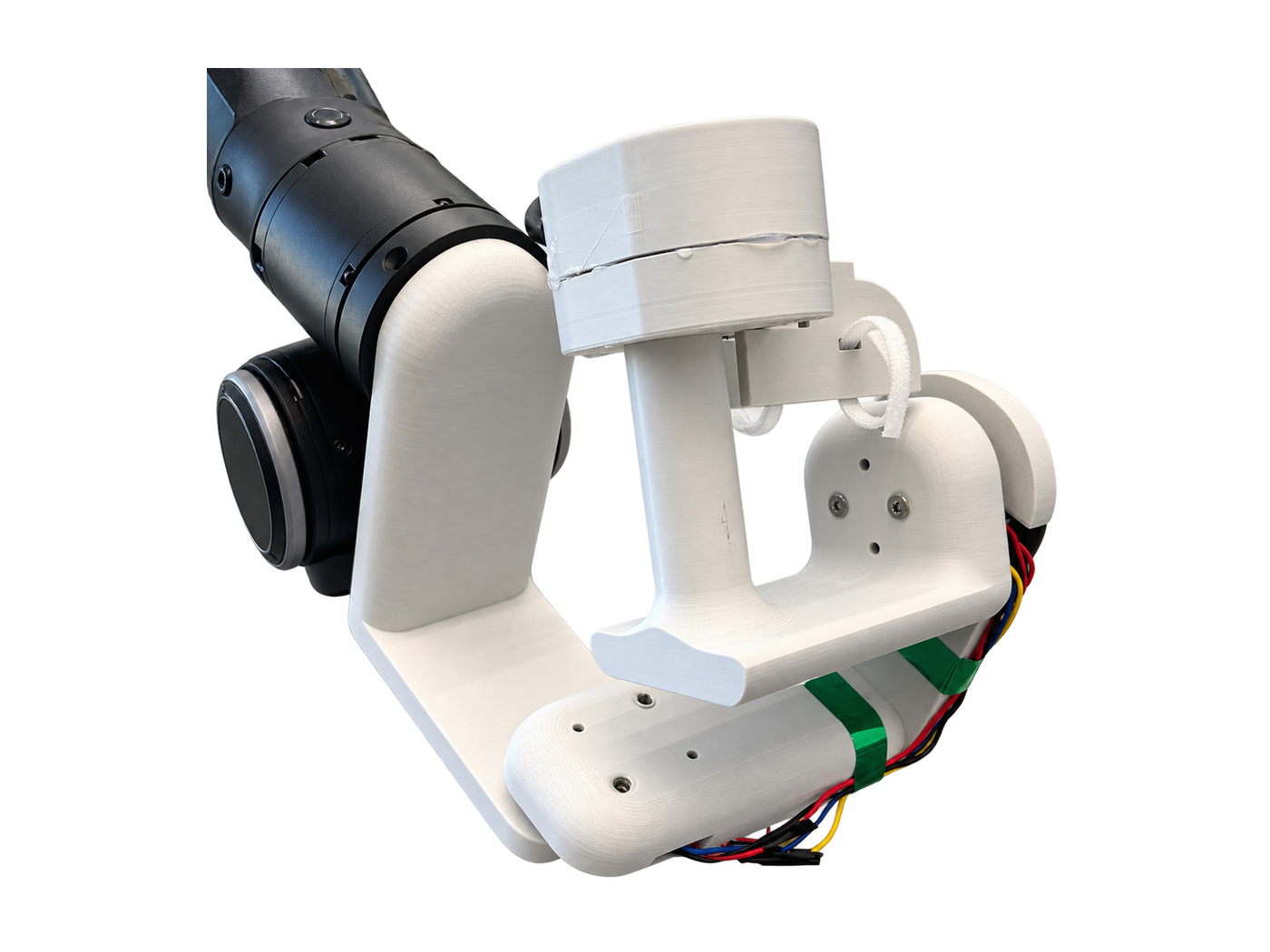}{(b)} &
    \panelh{\ph}{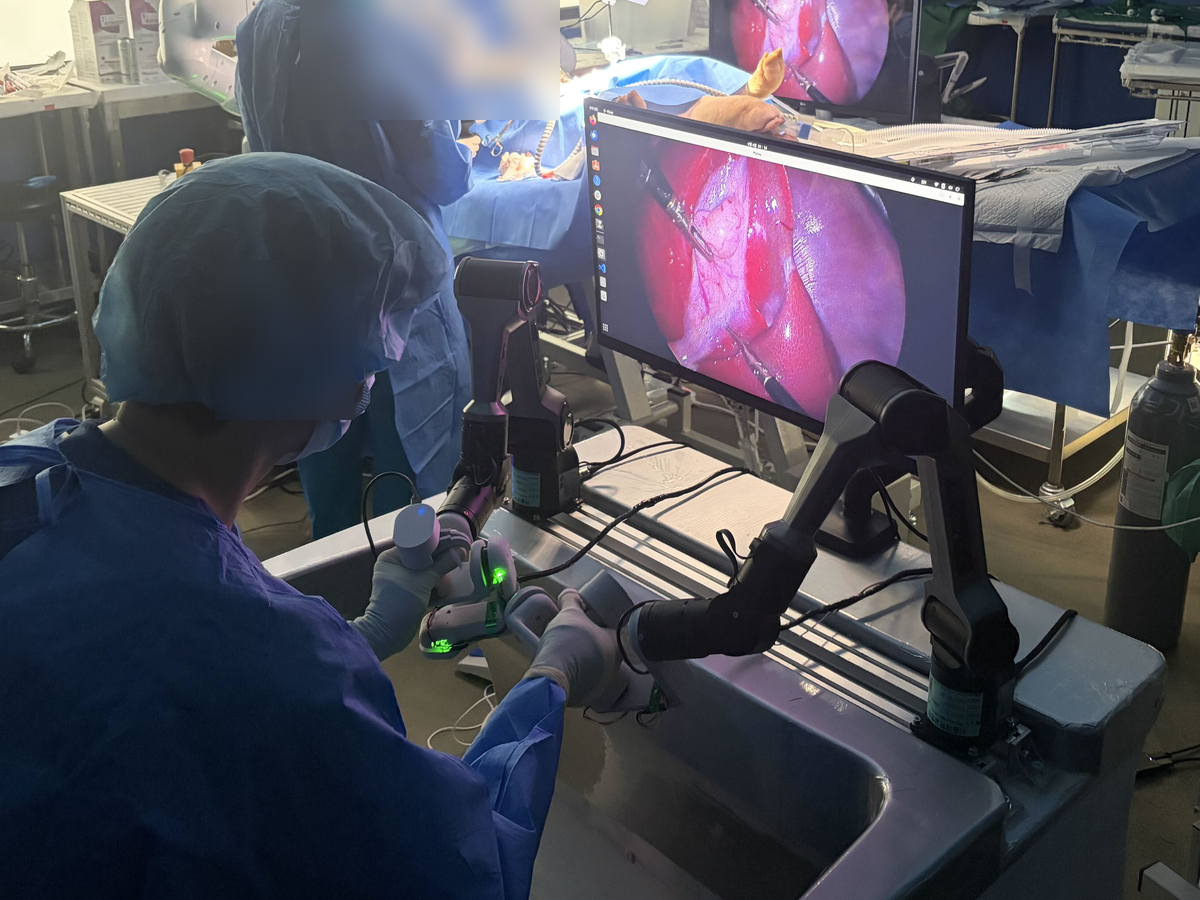}{(c)} &
    \paneltrimh{\ph}{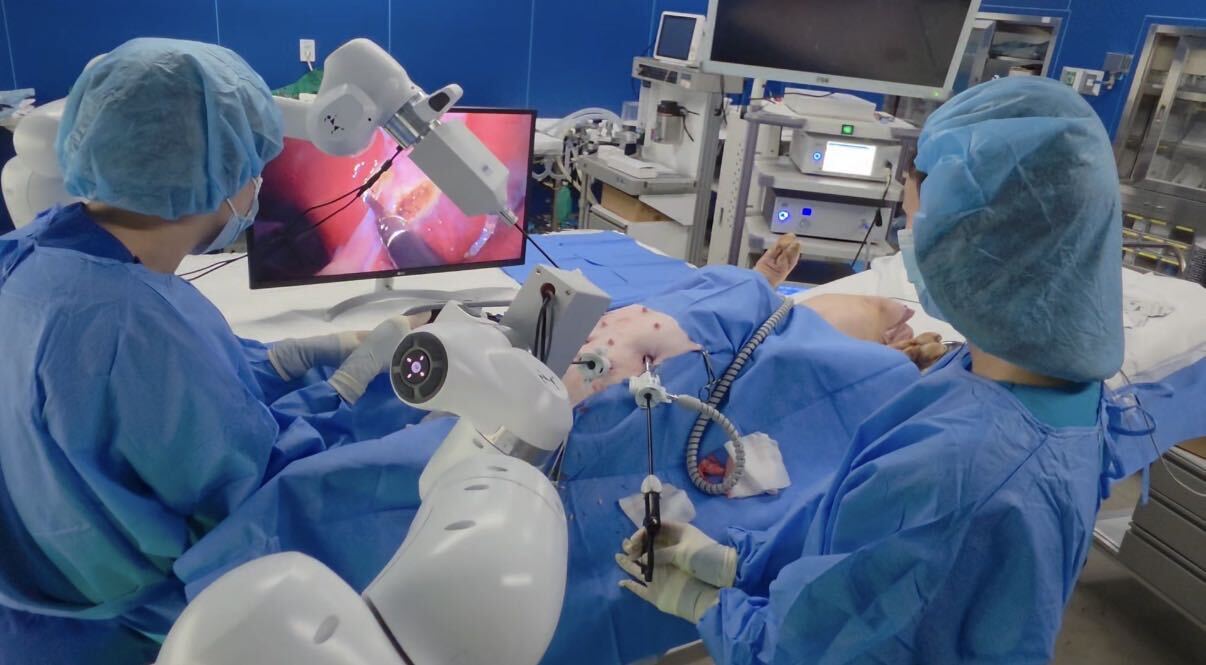}{(d)}{190 0 190 0} \\
  \end{tabular}}
  \captionof{figure}{\textbf{Physical implementation of the proposed surgical master and teleoperation setup.}
  (a) Bimanual master console built from two general-purpose 6-DoF robot arms.
  (b) The custom flange-mounted articulated distal adapter, which supplies the surgical-specific
  2-DoF wrist and continuous grasp interface absent from the general-purpose leader arm.
  (c) The same operator-side console used during the porcine experiment.
  (d) The pre-existing dual-FR3 patient-side laparoscopic platform; its 1-kHz RCM controller is unchanged.}
  \label{fig:hero}
\end{strip}

\begin{abstract}
High-performance surgical master consoles offer intuitive articulated manipulation but are expensive and difficult to reproduce, while accessible commercial haptic devices lack built-in interfaces for surgical wrist articulation and continuous grasp. We present a laparoscopic master in which general-purpose robot manipulators provide the programmable base and surgical-specific interaction is concentrated in a separable distal adapter. Each 6-DoF arm carries a custom 2-DoF direct-drive wrist, continuous grasp sensing, and clutch-based workspace management; the complete bimanual hardware costs approximately US\$5,500, below common bimanual commercial haptic references. To support sustained hand-guided use, we characterize current scaling, gravity, and break-away friction and use them for posture-dependent gravity compensation and motion-gated assistance, reducing continuous actuation and thermal loading. A common master state drives either Isaac Sim or an RCM-constrained dual-FR3 platform. Master-state publication to FR3 target update adds at most 8~ms, and tip tracking shows 0.31--0.66~mm mean delay-aligned residuals. In a porcine cholecystectomy experiment, five common operator-side metrics and hand/tip-speed distributions were compared descriptively with human in-vivo telemetry from 120 clinical Versius recordings of 99 cholecystectomy procedures; no interruption originated from the master console, and no abnormal clinical signs were recorded during the 7-day postoperative observation period. These results demonstrate a surgical master built from general-purpose robot arms with surgical-specific interaction concentrated in a separable distal interface.
\end{abstract}

\section{INTRODUCTION}
Robot-assisted minimally invasive surgery depends on an operator interface that can command wrist articulation and grasp while supporting workspace management over sustained manipulation. Dedicated surgical masters provide these functions but rely on specialized hardware that is difficult and costly to reproduce in a research setting \cite{kazanzides2014,zhang2020hamlyn}. Commercial stylus-type haptic devices are easier to obtain and have been used as master interfaces in laparoscopic teleoperation research \cite{berkelman2009,munawar2016}. Affordable simulator consoles have likewise paired commercial haptic devices with add-on grippers and pedals, but expert users still report differences in hand interface and available motion range relative to dedicated surgical consoles \cite{neri2024}. The distal mechanics supplied with these devices therefore do not directly reproduce an articulated surgical wrist with continuously variable finger grasp. Outside surgery, robot-like leaders such as ALOHA, GELLO, and FACTR show that general-purpose or replica manipulators can serve as intuitive masters when leader and follower kinematics are closely matched \cite{zhao2023aloha,wu2024gello,liu2025factr}. These lines of work leave a practical gap: surgical-specific hand interfaces are typically tied to dedicated or device-fixed master hardware, whereas robot-like leaders generally exploit kinematic correspondence with the follower. Using a general-purpose active arm with a non-isomorphic surgical follower therefore requires both hand-guided dynamic compensation and an operator representation decoupled from follower joint coordinates.

Rather than designing another purpose-built proximal master mechanism, we use a general-purpose research manipulator as the programmable master base and add a surgical hand interface through a separable distal module: a custom 2-DoF direct-drive wrist, continuous grasp sensing, and clutch input on the flange. This separation is intended to simplify hand-interface modification without redesigning the complete master; we evaluate one laparoscopic implementation, not cross-procedure adaptation. The resulting master drives followers that are kinematically non-isomorphic to it. Making the robot-arm base and distal module suitable for direct hand guidance requires treating their physical nonidealities explicitly: payload-dependent gravity, current scaling, break-away friction, and thermal loading are characterized and incorporated into the hand-guided controller.

The master exports tool pose, distal articulation, grasp, and clutch rather than patient-side joint targets. The same operator interface therefore drives either Isaac Sim or a pre-existing RCM-constrained dual-FR3 platform through endpoint-specific adapters.

The main contributions are:
\begin{enumerate}
    \item \textbf{A separable surgical-master design using a general-purpose manipulator as the programmable base:} commercially available 6-DoF arms provide the proximal base, while surgical-specific interaction is concentrated in a flange-mounted 2-DoF direct-drive wrist with continuous grasp and clutch input. This structure is intended to support hand-interface modification without redesigning the full proximal mechanism. The present work demonstrates one laparoscopic implementation; multi-interface and cross-vendor reconfiguration remain future work. The bimanual hardware costs approximately US\$5,500; CAD models, URDFs, and mechanical design files are available in the online repository to support reproduction and modification of the master hardware: \mechanicalrepo.
    \item \textbf{Hand-guided control and an endpoint-independent master-state interface for a non-isomorphic surgical follower:} external current/torque calibration and bidirectional break-away characterization separate posture-dependent gravity from friction and inform motion-gated assistance for sustained hand-guided use. The master exports tool pose, distal articulation, grasp, and clutch rather than follower joint targets, allowing the same state to drive a cholecystectomy-oriented Isaac Sim environment and an RCM-constrained dual-FR3 platform through endpoint-specific adapters.
    \item \textbf{End-to-end validation from bench characterization to porcine in-vivo use:} bench experiments quantify wrist behavior, tip tracking, clutch re-anchoring, and communication timing before the same physical console is used for simulator familiarization and a porcine cholecystectomy experiment. After 18~min of familiarization, the operator used the console throughout the 32.5-min robotic portion without any interruption originating from the master console; five common operator-side metrics and hand/tip-speed distributions were placed in descriptive context against 120 human in-vivo Versius recordings, and no abnormal clinical signs were recorded during 7-day postoperative observation.
\end{enumerate}

\section{RELATED WORK}

\subsection{Surgical Master Interfaces}
Dedicated systems such as the da Vinci/dVRK and Hamlyn CRM use surgical-specific master hardware \cite{kazanzides2014,zhang2020hamlyn}, while commercial haptic devices have been adapted for laparoscopic teleoperation \cite{berkelman2009,munawar2016,tobergte2011sigma7}. Neri \emph{et al.} added grippers and pedals to two Geomagic Touch devices in an affordable training console \cite{neri2024}. Prior work also emphasizes low interaction effort, ergonomic workspace design, and clutch-based recentering \cite{du2021hri,kang2022workspace,wong2023ergonomics,wong2024manipulation}. These systems motivate our design, but their distal mechanics remain dedicated or device-fixed. A stylus-type device supplies a pose and a single trigger, so wrist articulation must be inferred from the stylus orientation and grasp reduces to one aggregate opening; adding an articulated wrist or a separately sensed grip requires modifying hardware the vendor does not expose. Dedicated consoles provide these functions as a fixed assembly, leaving no incremental path to a different hand interface. Table~\ref{tab:related} summarizes the positioning.

\subsection{Kinematically Matched Robot-Like Leaders}
ALOHA, GELLO, and FACTR use matched or equivalent leader--follower kinematics for intuitive manipulation \cite{zhao2023aloha,wu2024gello,liu2025factr}; joint-mapped exoskeletons follow the same principle, while pose-retargeting systems relax it \cite{fang2024airexo,yang2024ace}. In our setting joint correspondence is unavailable by construction. The trocar fixes a remote centre of motion, so the shaft pitch and yaw of the follower are set by the port rather than by any leader joint, and a general-purpose arm carries no distal joint corresponding to the instrument wrist or jaw. Leader and follower are therefore non-isomorphic, operator intent is exported as tool pose, distal articulation, grasp, and clutch rather than as mirrored joint coordinates, and the missing surgical degrees of freedom are supplied by the distal adapter.

\subsection{RCM-Constrained Surgical Teleoperation}
Laparoscopic instruments passing through a trocar must respect the RCM constraint, and prior work combines follower-side enforcement with master-side haptic or hierarchical control: Marinho \emph{et al.} formulated the constraint as a follower-side optimization with master-side impedance feedback, and Su \emph{et al.} controlled RCM and surgical-tip motion hierarchically from a commercial haptic master \cite{marinho2019,su2020}. In our system the 1-kHz follower-side RCM controller is pre-existing and unchanged; the contribution lies upstream in the active master, re-anchoring, and endpoint-decoupled operator state.

\begin{table*}[!t]
\vspace*{1.5mm}
\caption{Positioning relative to representative teleoperation approaches.}
\label{tab:related}
\centering
\scriptsize
\setlength{\tabcolsep}{3.0pt}
\renewcommand{\arraystretch}{1.08}
\begin{tabular}{@{}p{0.175\textwidth}p{0.155\textwidth}p{0.205\textwidth}p{0.205\textwidth}p{0.205\textwidth}@{}}
\toprule
System(s) & Leader hardware & Leader--follower relation & Distal input / grasp & Reported validation \\
\midrule
\textbf{dVRK / Hamlyn CRM} \cite{kazanzides2014,zhang2020hamlyn}
& Dedicated surgical master
& Purpose-built; matched to the surgical platform by design
& Integrated wrist and grasp controls
& Open research platform; bench and ergonomic evaluation \\

\textbf{ALOHA / GELLO / FACTR} \cite{zhao2023aloha,wu2024gello,liu2025factr}
& Robot-like, replica, or commodity leader
& Kinematically matched or equivalent; joint correspondence exploited
& Lever, trigger, or platform-specific distal input
& Bimanual and contact-rich bench manipulation \\

\textbf{Marinho \emph{et al.} / Su \emph{et al.}} \cite{marinho2019,su2020}
& Commercial haptic device
& Non-isomorphic; RCM-constrained surgical follower
& Device-fixed stylus / gripper input
& Constrained surgical teleoperation \\

\textbf{TELESIM / Rodriguez \emph{et al.}} \cite{audonnet2024,rodriguez2026}
& VR, finger-mapping, or platform-dependent input
& Heterogeneous devices/endpoints; adapter or robot-agnostic integration
& Input-device dependent
& User study; phantom, ex-vivo, and in-vivo validation \\

\textbf{This work}
& General-purpose 6-DoF robot arm
& Non-isomorphic RCM follower; no patient-side joint target in the master message
& Custom 2-DoF wrist, continuous magnetic grasp, and clutch
& Bench and porcine in-vivo evaluation; clinical Versius reference comparison \\
\bottomrule
\end{tabular}

\par\vspace{1.5mm}
\centering
\begin{minipage}[t]{0.485\textwidth}
\caption{Approximate bimanual console hardware cost, excluding display, workstation, labor, and patient side. Listed reference devices are included for cost context only.}
\label{tab:cost}
\centering
\scriptsize
\setlength{\tabcolsep}{3.0pt}
\renewcommand{\arraystretch}{1.03}
\begin{tabular}{@{}lrr@{}}
\toprule
Item & Qty. & Cost (USD) \\
\midrule
PiPER 6-DoF arms & 2 & 3,998 \\
Direct-drive wrist actuators & 4 & $\approx$720 \\
Grasp sensors + distal MCUs & 2+2 & 40 \\
Printed distal parts / fasteners & -- & $\approx$60 \\
Foot pedals & 2 & $\approx$40 \\
Frame, arm supports, and mounts & 1 & $\approx$644 \\
\midrule
\textbf{Proposed console total} & & \textbf{$\approx$5,502} \\
\midrule
Phantom Touch, bimanual reference & 2 & 7,820 \\
sigma.7, bimanual reference & 2 & 96,600 \\
\bottomrule
\end{tabular}
\end{minipage}\hfill
\begin{minipage}[t]{0.485\textwidth}
\caption{Leader-arm parameters of Eq.~(\ref{eq:gravity}) used in the in-vivo session. $\mathbf{S}_g$ and $\mathbf{B}$ are diagonal; $\mathbf{S}_g$ is dimensionless, $\boldsymbol{\tau}_b$ is in N\,m, and $\mathbf{B}$ is in N\,m\,s/rad. Both arms run at 200~Hz with the payload-inclusive URDF. Distal wrist: $K_t=0.0613$~N\,m/A, 200-Hz loop, current limit 1.5~A per axis.}
\label{tab:params}
\centering
\scriptsize
\setlength{\tabcolsep}{3.2pt}
\renewcommand{\arraystretch}{1.03}
\begin{tabular}{@{}llrrrrrr@{}}
\toprule
 & Arm & J1 & J2 & J3 & J4 & J5 & J6 \\
\midrule
\multirow{2}{*}{$\mathbf{S}_g$} & R & 1.00 & 1.15 & 1.40 & 1.78 & 2.20 & 2.00 \\
 & L & 1.00 & 1.15 & 1.45 & 1.80 & 2.40 & 1.00 \\
\midrule
\multirow{2}{*}{$\boldsymbol{\tau}_b$} & R & 0 & $-0.10$ & $-0.05$ & 0 & $-0.20$ & 0.05 \\
 & L & 0 & $-0.20$ & 0 & 0.10 & $-0.20$ & $-0.05$ \\
\midrule
\multirow{2}{*}{$\mathbf{B}$} & R & 0 & 0 & 0 & 0.01 & 0 & 0.03 \\
 & L & 0 & 0 & 0 & 0.07 & 0 & 0.03 \\
\bottomrule
\end{tabular}
\end{minipage}
\end{table*}

\subsection{Modular Teleoperation and Surgical Research Platforms}
TELESIM and CRTK address heterogeneous robots through modular digital-twin or common robot-side interfaces \cite{audonnet2024,su2020crtk}, and Rodriguez \emph{et al.} integrated RCM control, teleoperation, data collection, and policy deployment on general-purpose manipulators \cite{rodriguez2026}. Our focus is complementary: the common abstraction sits on the operator side, coupled to a separable surgical hand interface, while mapping, safety, and RCM enforcement stay endpoint-local. Those platforms standardize the robot side so that different followers can be reached through one interface; here the invariant is the operator-side message, so a change of hand interface does not propagate into the follower stack and a change of follower does not require reworking the console.

\section{METHODS}

\subsection{System Overview}
\begin{figure*}[!t]
  \centering
  \includegraphics[width=0.90\textwidth,height=0.98\textheight,   keepaspectratio
]{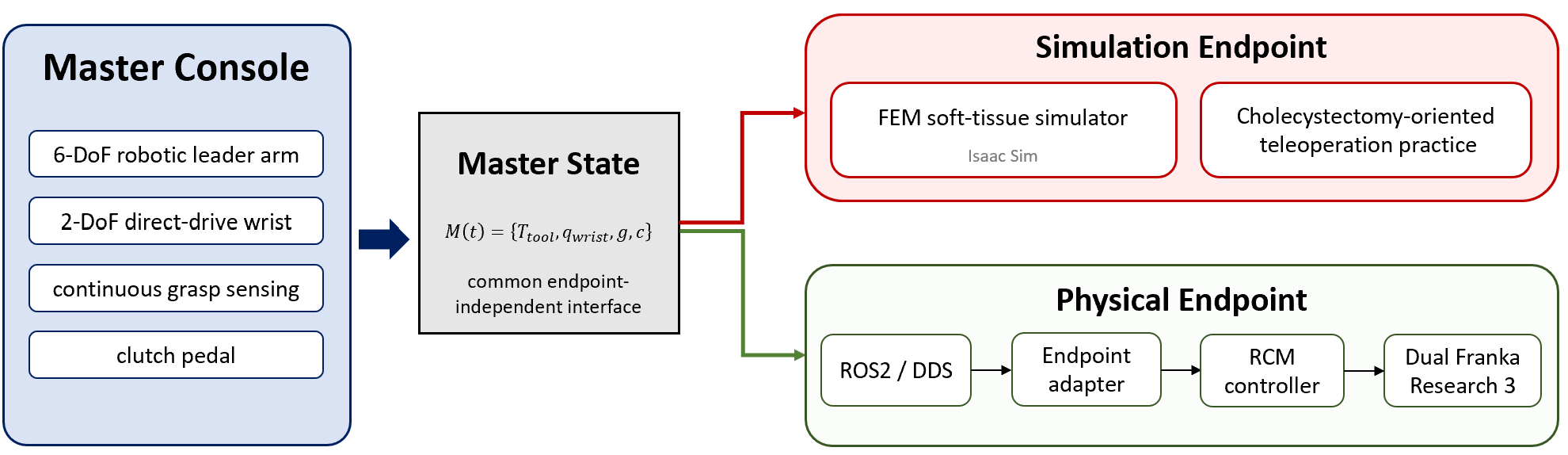}
  \caption{\textbf{System architecture and common endpoint-independent master state.}
  The master console combines a general-purpose 6-DoF robotic leader arm, the custom 2-DoF direct-drive wrist, continuous grasp sensing, and clutch input. These measurements are normalized into a common master state $\mathcal{M}(t)=\{T_{\mathrm{tool}},q_{\mathrm{wrist}},g,c\}$. The same state drives either (top) the Isaac Sim branch with FEM soft-tissue organs for cholecystectomy-oriented teleoperation practice or (bottom) the physical branch, where ROS~2/DDS and an endpoint-specific adapter map the operator command to the Franka Research 3 (FR3) laparoscopic platform. Endpoint-specific mapping, safety checks, and the pre-existing 1-kHz RCM controller remain on the patient side.}
  \label{fig:architecture}
\end{figure*}

Figures~\ref{fig:hero} and~\ref{fig:architecture} summarize the physical system and software boundary. The distal adapter is mechanically separable from the base arm, so adaptation to another leader is intended to require a robot-specific flange and updated kinematic/payload model rather than redesign of the hand interface; cross-vendor adaptation is not evaluated here.

\begin{figure*}[t]
  \centering
  \setlength{\tabcolsep}{2pt}
  \begin{tabular}{@{}cc@{}}
    \includegraphics[width=0.415\textwidth,height=0.175\textheight,keepaspectratio]{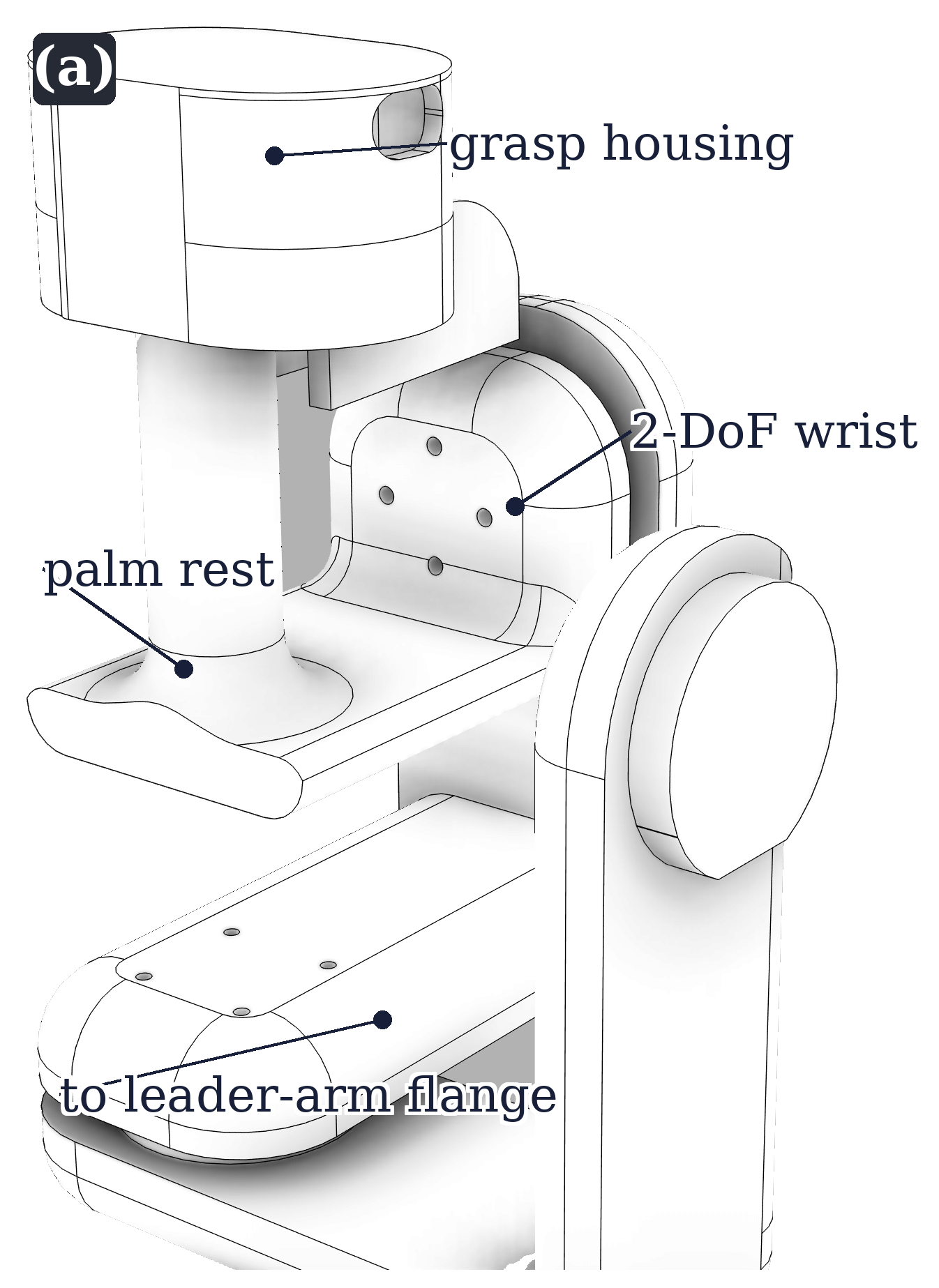} &
    \includegraphics[width=0.415\textwidth,height=0.175\textheight,keepaspectratio]{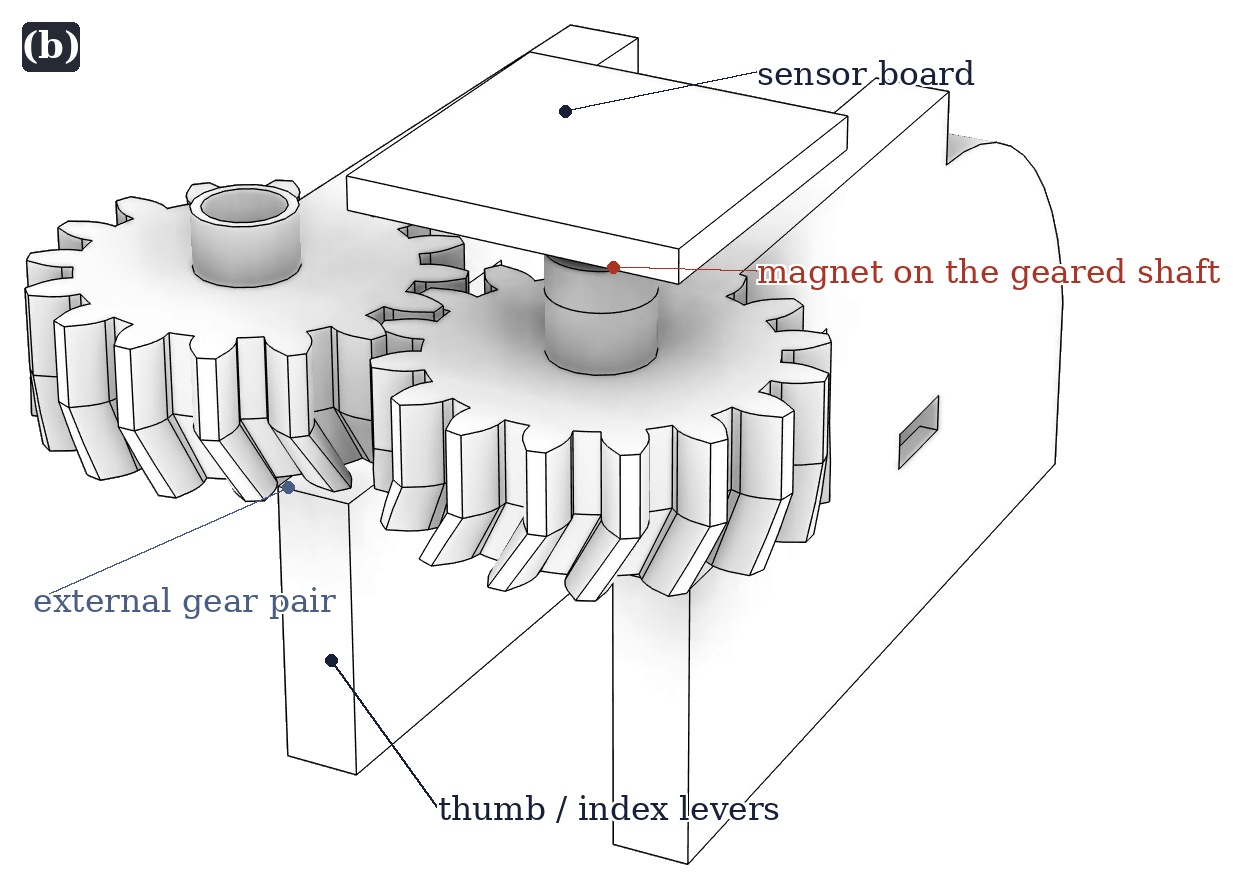} \\[2pt]
    \includegraphics[width=0.415\textwidth,height=0.175\textheight,keepaspectratio]{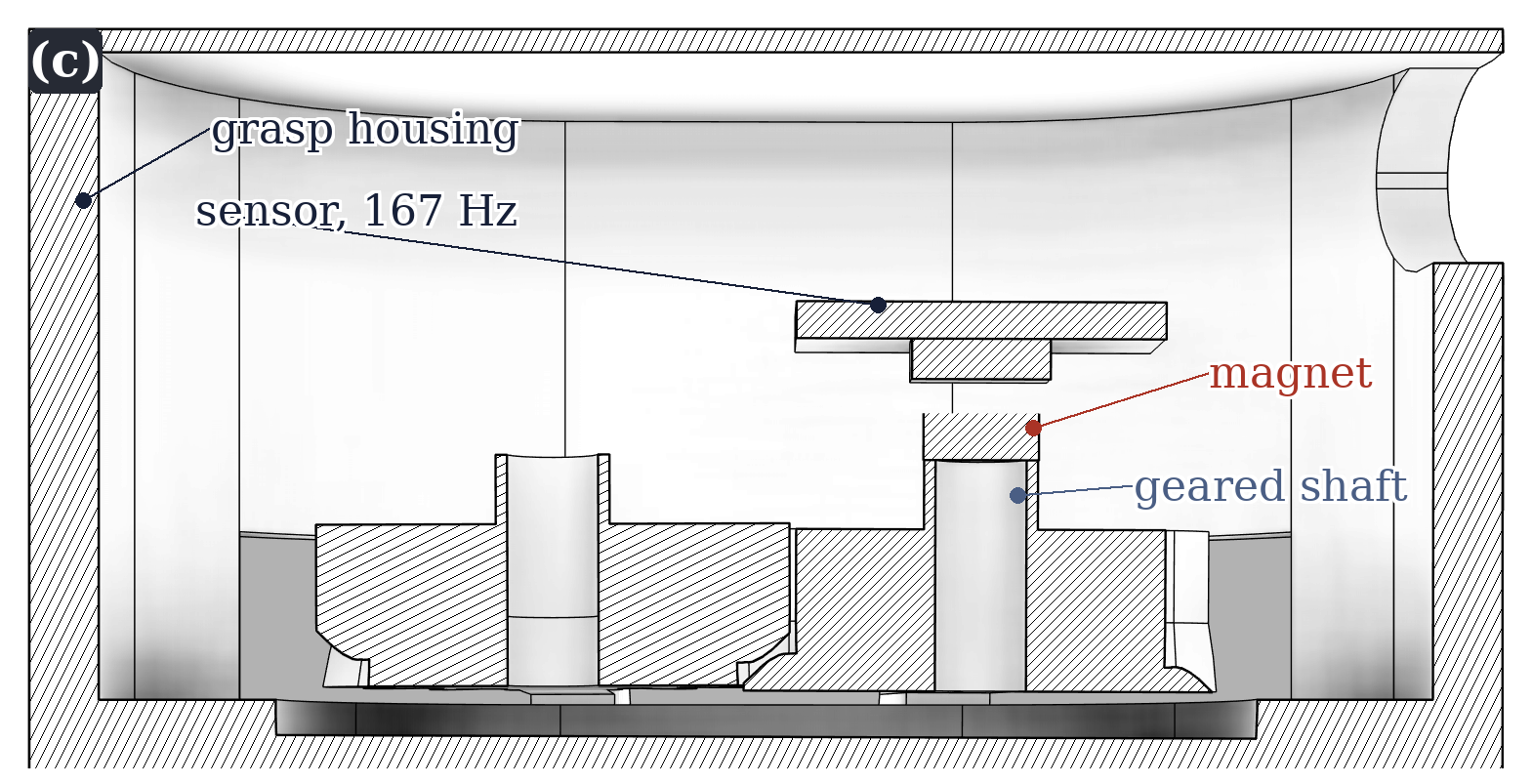} &
    \includegraphics[width=0.415\textwidth,height=0.175\textheight,keepaspectratio]{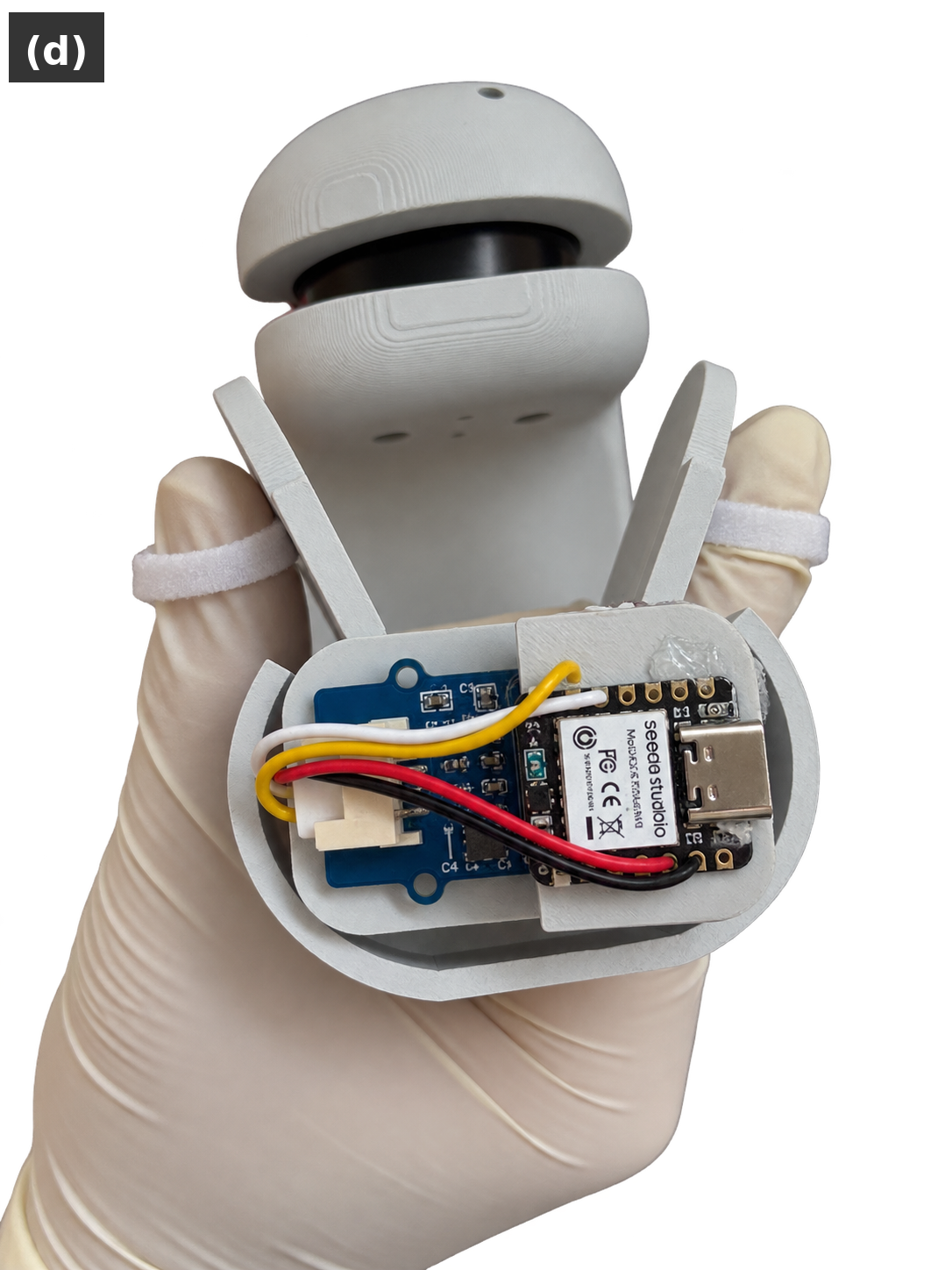}
  \end{tabular}
  \caption{\textbf{Mechanical design and implementation of the custom distal module.}
  (a) Overall flange-mounted module with palm rest and 2-DoF articulated wrist.
  (b) Passive thumb/index levers, external gear pair, geared-shaft magnet, and sensor board.
  (c) Section view of the sensor--magnet arrangement; grasp is sampled at 167~Hz.
  (d) Implemented electronics and operator hand interface.}
  \label{fig:distal_design}
\end{figure*}

\begin{figure*}[!t]
\vspace*{1.2mm}

  \centering
  \includegraphics[width=0.94\textwidth,height=0.98\textheight,   keepaspectratio
]{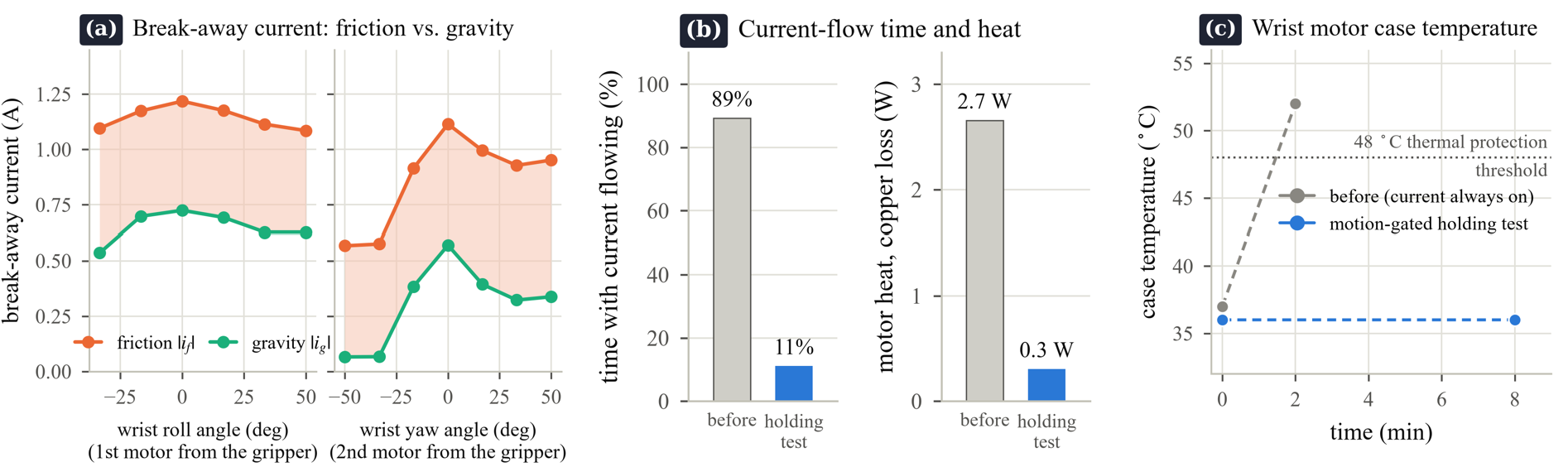}
  \caption{\textbf{Break-away characterization and sustained holding behavior of the direct-drive wrist.}
  (a) Friction and gravity currents separated by bidirectional break-away measurements over 13 roll/yaw wrist postures; friction exceeds gravity throughout the tested workspace.
  (b) Current-flow duty and estimated copper loss for the earlier current-on implementation and the final motion-gated holding test.
  (c) Motor-case temperature observations: the earlier current-on baseline rose from 37$^{\circ}$C to 52$^{\circ}$C within 2~min, whereas the final holding test remained near 36$^{\circ}$C over 8~min. The 48$^{\circ}$C dashed line denotes the wrist thermal-protection threshold used to prevent overheating, not an external contact-temperature standard.}
  \label{fig:wrist}
\end{figure*}

\begin{figure*}[!t]
  \centering
  \setlength{\tabcolsep}{2pt}
  \setlength{\ph}{46mm}
  \resizebox{0.95\textwidth}{!}{%
  \begin{tabular}{@{}cc@{}}
    \panelh{\ph}{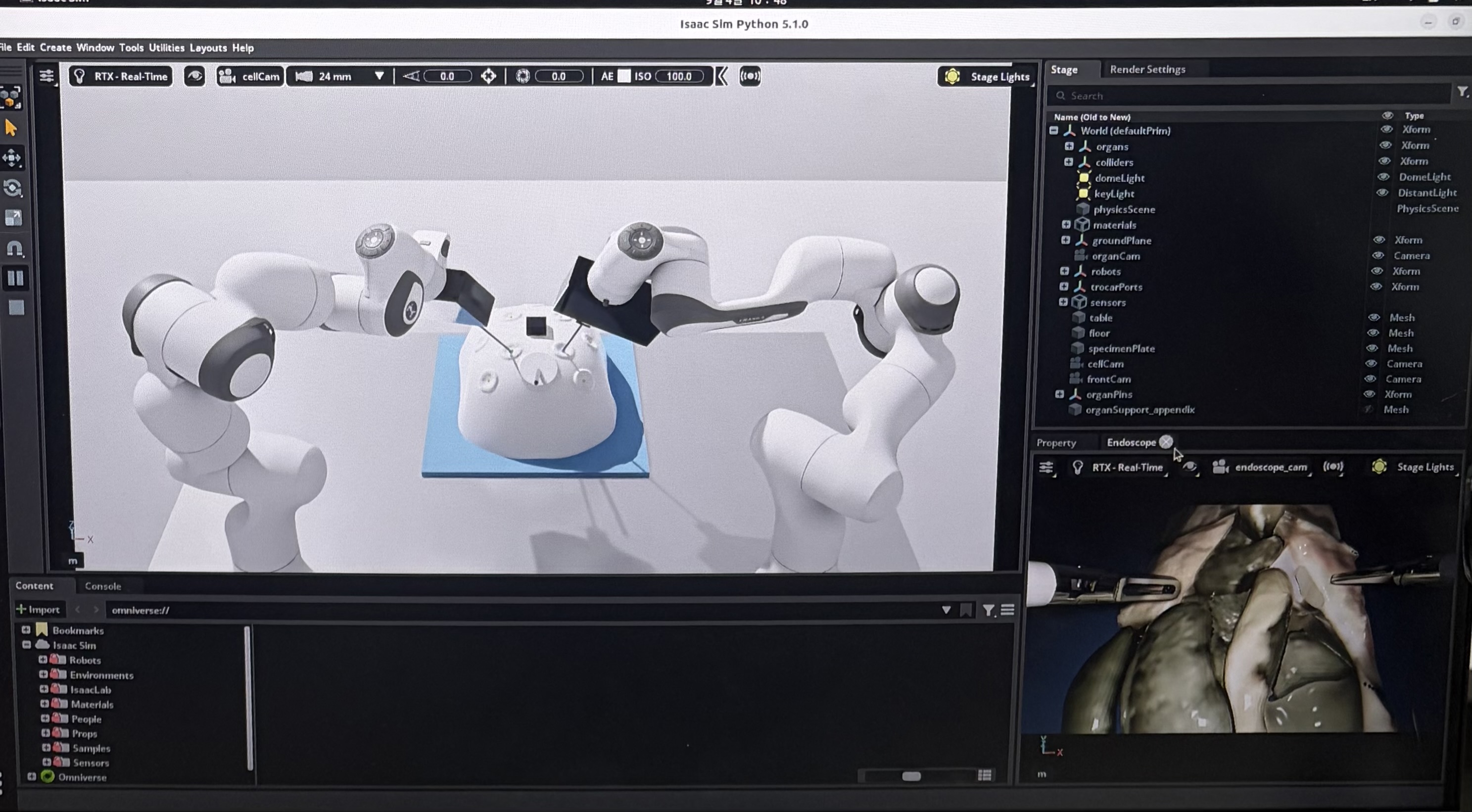}{(a)} &
    \panelh{\ph}{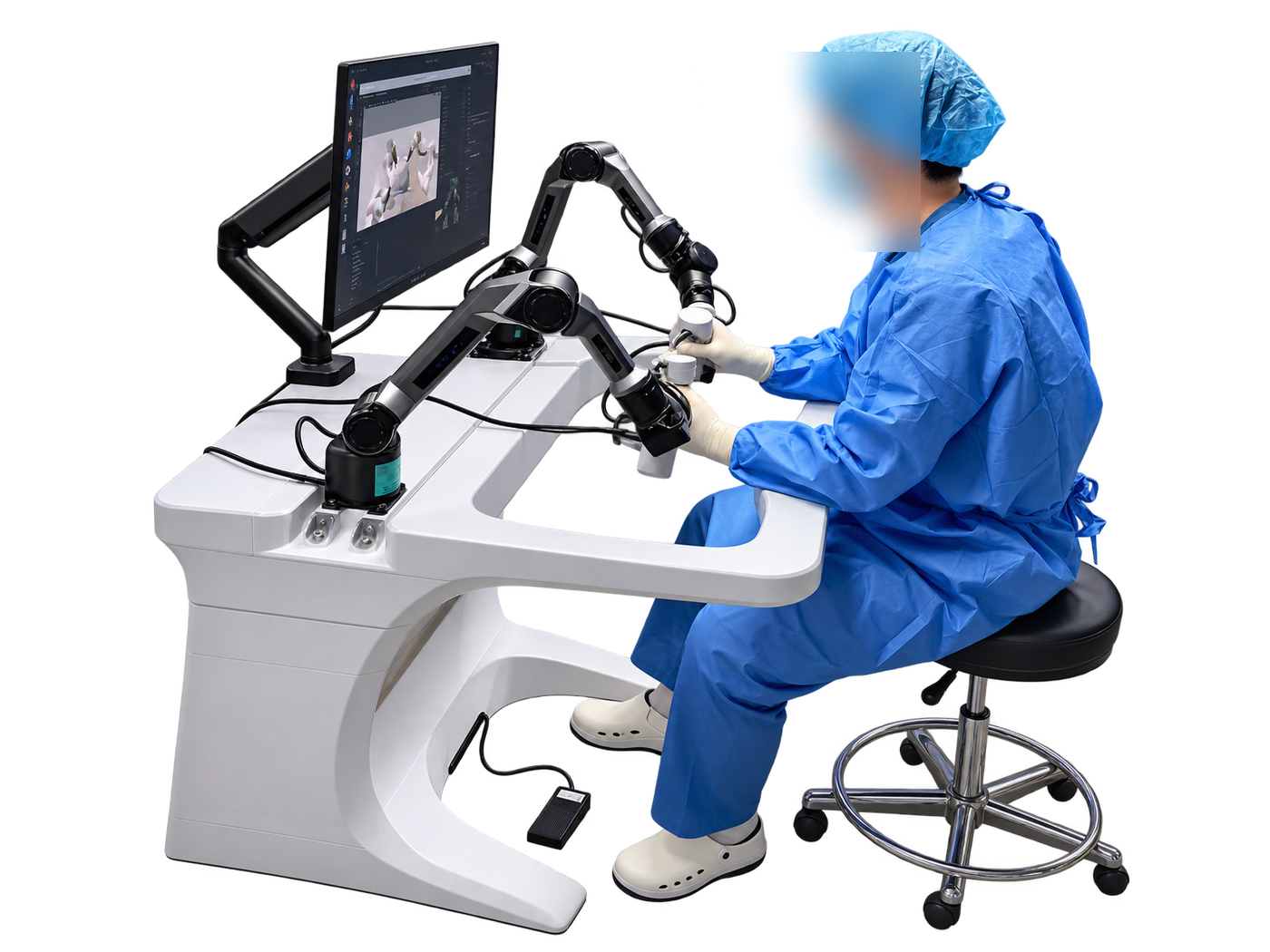}{(b)} \\
  \end{tabular}}
  \caption{\textbf{Isaac Sim environment and operator familiarization on the same console.}
  (a) The simulated dual-FR3 laparoscopic setup operating through fixed trocar ports in a phantom
  containing a simulated gallbladder model, with the endoscopic view at lower right. (b) The veterinarian practicing in the simulator through the same
  physical master console later used in the porcine experiment; no operator-side hardware or
  mapping was changed between the two settings.}
  \label{fig:simulation}
\end{figure*}

\subsection{Robot-Arm-Based Active Master}
The base arms are controlled from the host computer at 200~Hz rather than through a firmware teaching mode. Joint positions and velocities are read from the arm and torque commands are generated as
\begin{equation}
\boldsymbol{\tau}_{m}=\mathbf{S}_{g}\mathbf{g}(\mathbf{q})+\boldsymbol{\tau}_{b}-\mathbf{B}\dot{\mathbf{q}},
\label{eq:gravity}
\end{equation}
where $\mathbf{g}(\mathbf{q})$ is model-predicted gravity torque, $\mathbf{S}_{g}$ is a joint-wise scale, $\boldsymbol{\tau}_{b}$ is residual trim, and $\mathbf{B}$ provides small viscous stabilization. The stock end-effector was removed and replaced by the custom distal module, so the manufacturer's nominal distal-link inertial model no longer exactly represents the assembled leader. Residual modeling and drive-side errors were therefore compensated by joint-wise gravity scale factors, which increased from J1 to J5 in both arms (Table~\ref{tab:params}). Left and right parameters were calibrated independently to account for unit-specific drive and friction characteristics; the asymmetry in Table~\ref{tab:params} reflects this per-unit calibration rather than a difference in the mounted distal module. Because the drive is commanded with zero position gain and a desired velocity equal to the measured velocity, its internal damping term cancels, so $\mathbf{B}$ is applied on the host. Table~\ref{tab:params} lists the values used in the in-vivo session. The arms serve as gravity-compensated input devices; no force feedback is rendered to the operator.

\subsection{Distal Module Design and Continuous Grasp}
The distal module adds two orthogonal direct-drive rotational axes, giving $\mathbf{q}_{m}=[q_1,\ldots,q_6,q_7,q_8]^T$. Figure~\ref{fig:distal_design} shows the overall module, geared grasp transmission, sensor--magnet section, and implemented hand interface. Passive thumb/index levers drive a geared shaft carrying a permanent magnet; a magnetic rotary sensor measures shaft angle at 167~Hz across a fixed air gap. Grasp is therefore a passive continuously sensed scalar $g\in[0,1]$, not an additional kinematic DoF, and is mapped to endpoint jaw aperture. Both wrist axes are fully articulated and exported in $\mathcal{M}$, letting the operator reorient and twist the handle without moving the base chain. The rigid instrument on the present FR3 follower has no distal wrist, so that endpoint uses only the resulting twist as roll, whereas an articulated-instrument endpoint can map $\mathbf{q}_{w}$ directly.

\subsection{Direct-Drive Wrist Identification and Compensation}
The reducer-free wrist is backdrivable \cite{wensing2017proprioceptive} and directly manipulated by the operator, so actuator calibration and break-away behavior must be identified externally. Four commanded current levels from 0.30 to 1.50~A produced a drive-reported to commanded current ratio of $0.631\pm0.022$. A 4-g load at a 50-mm lever arm gave $K_t=0.0613$~N$\,$m/A from bidirectional break-away averages before and after loading, consistent with the manufacturer's 0.05--0.07~N$\,$m/A range; subsequent current results use the corrected scale. Bidirectional break-away sweeps over 13 wrist configurations separate gravity and friction as
\begin{equation}
 i_g=\frac{i^{+}+i^{-}}{2}, \qquad i_f=\frac{i^{+}-i^{-}}{2}.
\label{eq:wrist_sep}
\end{equation}
The resulting controller combines posture-dependent gravity compensation with friction assistance gated on detected operator motion, allowing output to fall near zero at rest instead of applying a fixed current continuously.

\subsection{Endpoint-Independent Interface and Mapping}
Operator intent is normalized into
\begin{equation}
\mathcal{M}(t)=\left\{{}^{W}\mathbf{T}_{\mathrm{tool}}(t),\mathbf{q}_{w}(t),g(t),c(t)\right\},
\label{eq:masterstate}
\end{equation}
where $\mathbf{q}_{w}=[q_7,q_8]^T$, $g$ is grasp, and $c$ is clutch state. Isaac Sim consumes the state through the simulation endpoint, whereas the physical path serializes the same state as a 96-B ROS~2 message at 200~Hz over DDS and dedicated Ethernet. On the physical branch, the endpoint adapter applies frame conversion, scaling, validity checks, and clutch anchoring before passing the tool target to the existing RCM controller.

When the clutch is engaged at $t_0$, master and follower poses are stored as anchors and translation is mapped as
\begin{equation}
\mathbf{p}^{d}_{s}(t)=\mathbf{p}_{s,0}+s\mathbf{A}[\mathbf{p}_{m}(t)-\mathbf{p}_{m,0}],
\label{eq:mapping}
\end{equation}
with $s=1/3$ in all experiments, while orientation contributes only its incremental twist about the tool axis, sent as instrument roll; shaft pitch and yaw are set by the trocar RCM. A bounded leader workspace is conventionally extended by indexing \cite{conti2005spanning}; while released, the follower target tracks the measured tip, so re-engagement re-establishes the master--follower offset at the instrument's current pose. Mapping, command validation, safety authority, and RCM enforcement remain endpoint-local.

\subsection{Simulation Environment}
For familiarization and endpoint verification, the same master-state interface drove an Isaac Sim scene with a rabbit gallbladder phantom and dual FR3 models constrained by fixed trocar ports (Fig.~\ref{fig:simulation}).

\section{EXPERIMENTS}


\subsection{Bench Hardware and Teleoperation Tests}
Current-flow duty, estimated copper loss, and motor-case temperature were compared between the earlier current-on implementation and the final motion-gated controller. The continuous grasp channel was checked against the simulated jaw command. Tip tracking used repeated line, circular, and free 3-D motions of approximately 40~s each; the error is the distance between the commanded tip target sent to the RCM controller and the measured tip position (FR3 base frame), after removing the best-aligning delay, on samples moving faster than 2~mm/s. Four clutch trials measured master repositioning and tool motion across release/re-engagement.

\subsection{Communication and Integration Timing}
The 96-B ROS~2 master-state message was transmitted at 200~Hz for 30~s over the dedicated Ethernet link. Packet delivery and sender-clock RTT were measured independently of robot motion. We also measured dead time from master-state publication to FR3 target-command update, including transport, 200-Hz teleoperation-node scheduling, mapping, and target publication but excluding subsequent RCM-controller, actuator, and mechanical response.

\subsection{In-Vivo Protocol}
This single-case feasibility evaluation was not designed as a user study. A veterinarian received 18~min of familiarization in the Isaac Sim environment of Fig.~\ref{fig:simulation}, using the same physical master console, before a single porcine cholecystectomy experiment. The robotic portion focused on gallbladder manipulation and dissection; remaining steps were completed conventionally. The robotic interval lasted 32.5~min; a 3.7-min patient-side state-feedback stall requiring software restart was excluded from motion analysis, leaving 28.4~min of analyzable data.

For clinical context, we additionally analyzed publicly available telemetry from the CMR Surgical Versius subset of the Open-H-Embodiment dataset \cite{openhembodiment2026}. The release contains human in-vivo surgical recordings from the Versius system and provides the two surgeon hand-controller poses together with pince, clutch, engagement state, and translation/rotation scaling. The cholecystectomy subset comprises 99 procedures released as 124 recordings, of which 120 carried the channels required here; because several recordings originate from the same procedure, the reference distribution is treated as a pooled description rather than as 120 independent cases. We retained five scalar operator-side metrics that could be defined consistently from the available Versius telemetry and our porcine session, covering workspace management and grip/handle behavior; hand-speed and commanded-tip-speed distributions were analyzed separately. The clinical data are used only as a descriptive reference, not as a matched control or an equivalence benchmark. Re-anchoring continuity was evaluated over the 46 valid in-vivo releases outside the interruption interval. The animal underwent daily general-condition monitoring for 7~days after the procedure under BIOSTEP IACUC protocol S26-PZ-0563.

In-vivo tip tracking followed the bench definition (100~Hz, engaged samples). Roll error was the shaft-axis angle between the measured tool axis and the controller's parallel-transported roll reference, delay-aligned on samples with commanded roll rate above 5$^{\circ}$/s.


\section{RESULTS}

\begin{figure*}[!t]
\vspace*{1.2mm}

    \centering
    \includegraphics[width=0.78\textwidth,height=0.98\textheight,   keepaspectratio
]{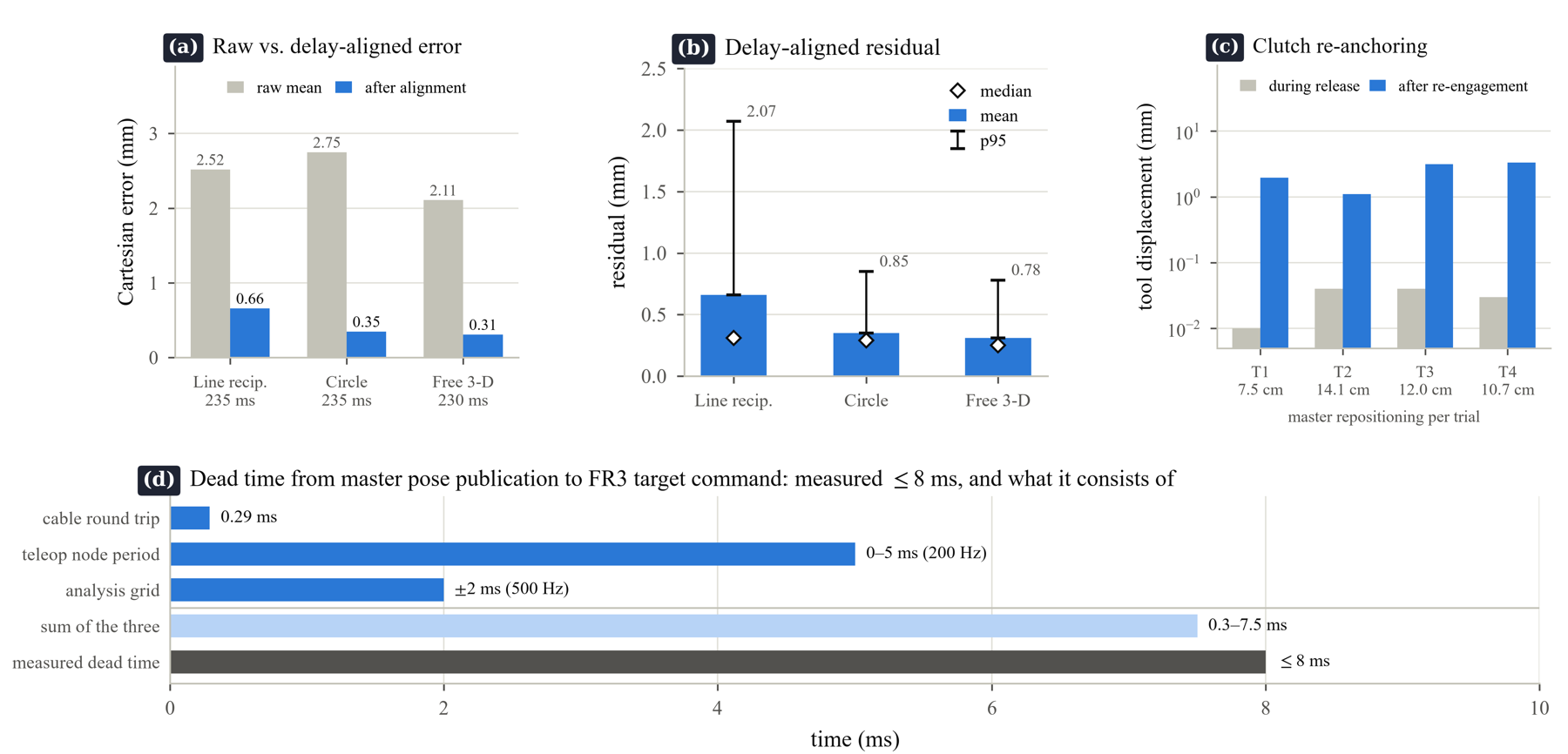}
    \caption{\textbf{Bench evaluation of the physical teleoperation path and master-to-target integration timing.}
    (a) Raw and delay-aligned Cartesian error between the commanded tool-tip target and the measured tip position (FR3 base frame; samples with commanded tip speed $>2$~mm/s) for line, circular, and free 3-D trajectories.
    (b) Delay-aligned residual statistics.
    (c) Clutch re-anchoring across four bench trials.
    (d) Measured timing from master-state publication to FR3 target-command update; the proposed path is $\leq 8$~ms and excludes patient-side RCM-controller and actuator response.}
    \label{fig:bench}
\end{figure*}

\begin{figure*}[!t]
  \centering
  \includegraphics[width=0.97\textwidth]{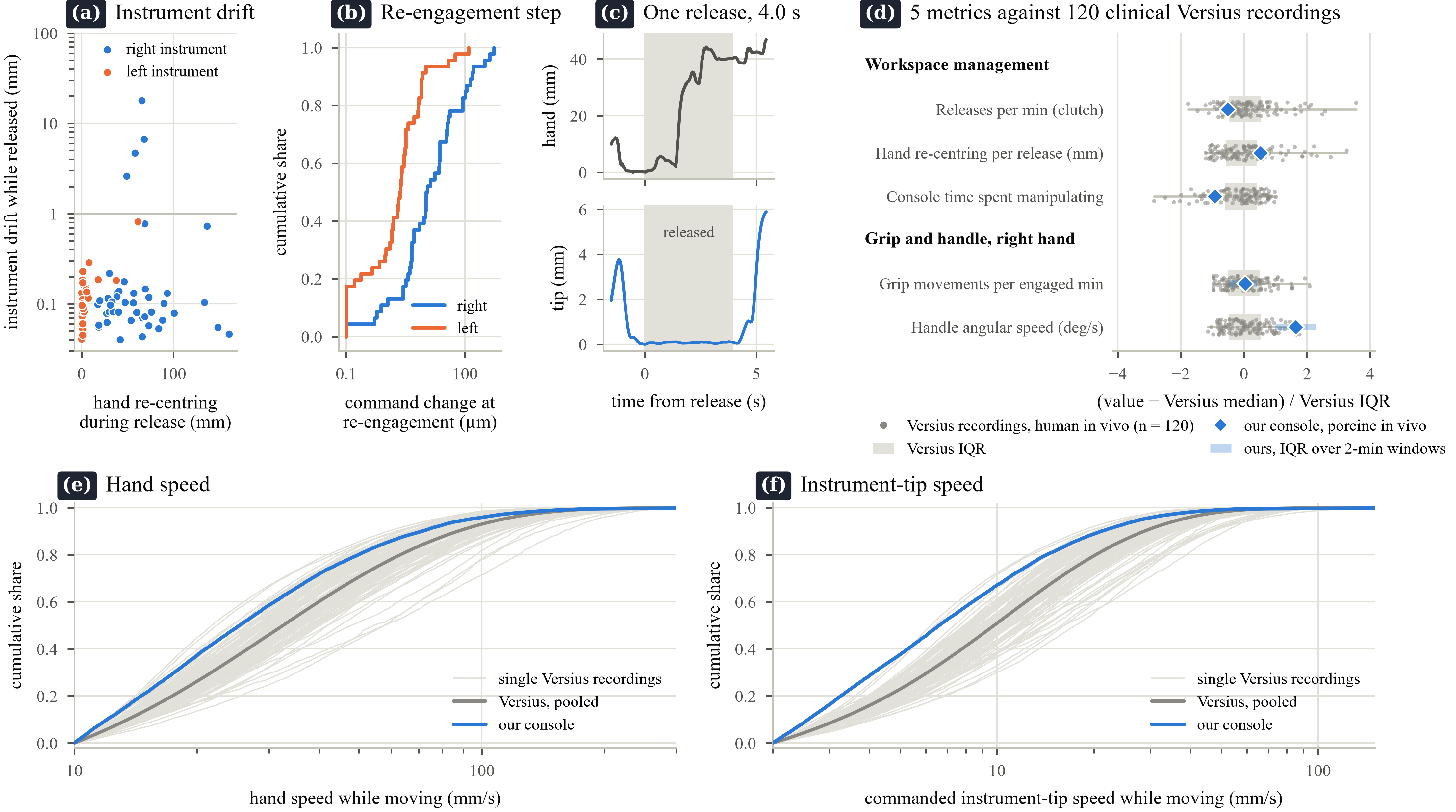}
  \caption{\textbf{Porcine in-vivo re-anchoring and clinical reference comparison.}
  (a) Instrument drift while the hand is re-centered during 46 valid releases.
  (b) Target-command discontinuity at re-engagement, in micrometres.
  (c) Representative release showing centimeter-scale hand motion while the instrument tip remains nearly stationary.
  (d) Five operator-side metrics---clutch releases per minute, hand re-centring per release, share of console time spent manipulating, grip movements per engaged minute, and handle angular speed---normalized by the median and IQR of 120 human in-vivo Versius recordings (99 cholecystectomy procedures); the blue diamond denotes the proposed console and the light-blue interval its IQR over 2-min windows.
  (e,f) Cumulative distributions of hand speed and commanded instrument-tip speed; thin gray curves show individual Versius recordings, the dark-gray curve pooled clinical data, and blue the proposed console. The Versius data provide descriptive clinical context rather than a matched benchmark.}
  \label{fig:invivo_compare}
\end{figure*}

\subsection{Distal-Module Hardware Characterization}
The break-away measurements in Fig.~\ref{fig:wrist}(a) showed that friction exceeded gravity at all 13 tested wrist postures, confirming that the reducer-free wrist is friction-dominated near rest.

This characterization directly informed the final holding strategy. Relative to the earlier implementation, current-flow duty fell from 89\% to 11\%, while estimated copper loss fell from 2.7 to 0.3~W (Fig.~\ref{fig:wrist}(b)). In separate temperature observations, the earlier current-on baseline rose from 37$^{\circ}$C to 52$^{\circ}$C within 2~min, whereas the final motion-gated holding test remained near 36$^{\circ}$C over 8~min (Fig.~\ref{fig:wrist}(c)).  In vivo, the grasp channel resolved one 12-bit sensor count, 0.45\% of the full right-hand opening, and showed no count change in 99.8\% of 1-s windows in which the lever was held within 2\% of full opening.

\subsection{Bench Teleoperation and Integration}
Figure~\ref{fig:bench} summarizes bench tip tracking, clutch continuity, and integration timing. After removing a 230--235~ms delay set by the pre-existing follower controller, mean errors between the commanded tip target and the measured tip fell from 2.11--2.75~mm to 0.31--0.66~mm; p95 was below 1~mm for circular and free-space motion and 2.07~mm for the reversing line trajectory. Across four clutch trials, 7.5--14.1~cm of master repositioning produced 0.01--0.04~mm tool motion during release and 1.11--3.34~mm immediately after re-engagement, as the instrument followed hand motion already under way when the pedal was pressed. All 6001 ROS~2 packets were received; mean RTT was 0.29~ms, with a p95 of 0.44~ms and a maximum of 2.46~ms. The measured dead time from master-state publication to FR3 target-command update was at most 8~ms (Fig.~\ref{fig:bench}(d)). The link therefore accounts for roughly 4\% of that bound, the remainder being dominated by the 200-Hz scheduling of the robot-side teleoperation node, so the transport is not the limiting term on this path. The bound is measured between the two published messages and excludes the downstream patient-side controller and mechanical response, which set the 230--235~ms delay removed above.

\subsection{Porcine In-Vivo Evaluation and Clinical Context}
No interruption originated from the master console: all 101 pedal transitions changed the engagement state within 0.2~s, and the only engagement change without a pedal command was the software stop at the end of the procedure. The single interruption inside the robotic interval was localized to the patient side: throughout the 3.7-min stall the master-state and target-command streams continued with no gap longer than 20~ms while both measured tip poses remained frozen.

Across 46 valid in-vivo re-anchorings, the right hand was repositioned by a median 56.5~mm while median instrument motion during release was 0.10~mm; 42 of 46 right-side releases stayed below 1~mm (Fig.~\ref{fig:invivo_compare}(a)). In the other four (2.6--17.8~mm) the instrument moved while released, which the tip-tracking target of Section~III-E follows without correction. Median target-command discontinuity at re-engagement was 11~$\mu$m (Fig.~\ref{fig:invivo_compare}(b)), and a representative release is shown in Fig.~\ref{fig:invivo_compare}(c).

Figure~\ref{fig:invivo_compare}(d--f) places the session in clinical context: all five operator-side metrics fell within the range spanned by the 120 Versius recordings. Grip movements per engaged minute was closest to the clinical median ($+0.04$~IQR); clutch releases per minute and hand re-centring per release lay at the edges of the interquartile range ($-0.51$ and $+0.53$~IQR); and two metrics fell outside it, with a lower share of console time spent manipulating ($-0.92$~IQR) and a higher handle angular speed ($+1.65$~IQR). The higher handle angular speed is consistent with the articulated distal adapter, whose wrist axes are rotated directly by the operator's hand; the lower manipulating-time share reflects a single porcine session. Hand speed overlapped the pooled clinical curve, while commanded tip speed was shifted slower, consistent with the smaller motion scale. The comparison is descriptive, not an equivalence test.

In vivo, removing a 230-ms delay reduced the median tip error from 1.28 to 0.28~mm (p95 3.7~mm) on the right and from 0.66 to 0.13~mm (p95 0.73~mm) on the left; the residual increased with tip speed. Right-hand roll, used actively (median 43$^{\circ}$ per engagement), showed a median residual of 0.32$^{\circ}$ (p95 6.7$^{\circ}$); left-hand roll was rarely used. The animal completed the 7-day postoperative observation period with no abnormal clinical signs recorded during daily general-condition monitoring.


\section{DISCUSSION AND LIMITATIONS}
The central design choice is to use a general-purpose robot arm as the master base and concentrate surgical-specific mechanics in a separable distal module. Unlike dedicated surgical masters, the surgical interface is thus an add-on to a commercially available arm, so other groups can build the console from off-the-shelf parts and adapt its hand interface or endpoints to their own research. The trade-off is that payload, friction, and heat become explicit design constraints; extending prior work on friction compensation and workspace design \cite{du2021hri,kang2022workspace} to a reducer-free add-on wrist, we measure these nonidealities and use them for low-duty hand-guided operation.

Unlike CRTK and digital-twin frameworks \cite{audonnet2024,su2020crtk}, the common abstraction sits on the operator side, so endpoint-specific safety and RCM control remain local. The $\leq8$~ms dead time therefore bounds only the master-side path; the $\sim$230-ms tip lag stems from the first-order response of the patient-side controller (identified time constant 203--205~ms). The master wrist provides two actively articulated DoFs, which were exercised during bench and simulator operation; the rigid laparoscopic follower used in vivo has no distal articulation, so only the component corresponding to instrument roll was mapped to the patient side. Orientation fidelity is therefore reported for roll only.

The Versius recordings provide clinical context rather than a matched control: our evidence is one porcine in-vivo case with one operator, whereas the reference data come from human clinical procedures performed by expert surgeons on a different robotic platform with its own scaling and control characteristics. Accordingly, overlap in the reported distributions should not be interpreted as equivalence between systems or as evidence of clinical readiness. Formal usability conclusions require multiple procedures, operators, and workload measures. A released instrument is not actively held against external load, and no force feedback is rendered; adding it would require tip sensing or estimation \cite{hosseinabadi2022force}. Leader-arm transparency was not quantified directly, so the clinical motion metrics should not be interpreted as a direct measure of back-driving effort.

\section{CONCLUSION}
We presented an accessible surgical master that combines general-purpose robot arms with a separable articulated distal adapter providing wrist articulation, continuous grasp input, and clutch-based workspace management. Hardware characterization separated gravity and friction and informed a low-duty hand-guided controller, while an endpoint-independent master state supported both simulation and an RCM-constrained physical platform. Bench and porcine in-vivo evaluation showed low-residual command transfer and stable re-anchoring, while common operator-side metrics and hand/tip-speed distributions were placed in descriptive context against human clinical Versius telemetry. The platform therefore provides an accessible and modifiable foundation for future laparoscopic teleoperation research.

\begingroup
\small
\setlength{\itemsep}{-1pt}

\endgroup


\begin{thebibliography}{99}

\bibitem{kazanzides2014}
P. Kazanzides, Z. Chen, A. Deguet, G. S. Fischer, R. H. Taylor, and S. P. DiMaio,
``An open-source research kit for the da Vinci surgical system,''
in \emph{Proc. IEEE Int. Conf. Robot. Autom. (ICRA)}, 2014, pp. 6434--6439.

\bibitem{zhang2020hamlyn}
D. Zhang, J. Liu, L. Zhang, and G.-Z. Yang,
``Hamlyn CRM: a compact master manipulator for surgical robot remote control,''
\emph{Int. J. Comput. Assist. Radiol. Surg.}, vol. 15, no. 3, pp. 503--514, 2020.

\bibitem{berkelman2009}
P. Berkelman and J. Ma,
``A compact modular teleoperated robotic system for laparoscopic surgery,''
\emph{Int. J. Robot. Res.}, vol. 28, no. 9, pp. 1198--1215, 2009.

\bibitem{munawar2016}
A. Munawar and G. Fischer,
``A surgical robot teleoperation framework for providing haptic feedback incorporating virtual environment-based guidance,''
\emph{Front. Robot. AI}, vol. 3, Art. no. 47, 2016.

\bibitem{neri2024}
A. Neri \emph{et al.},
``A novel affordable user interface for robotic surgery training: design, development and usability study,''
\emph{Front. Digit. Health}, vol. 6, Art. no. 1428534, 2024.

\bibitem{zhao2023aloha}
T. Z. Zhao, V. Kumar, S. Levine, and C. Finn,
``Learning fine-grained bimanual manipulation with low-cost hardware,''
in \emph{Proc. Robotics: Science and Systems (RSS)}, 2023.

\bibitem{wu2024gello}
P. Wu, Y. Shentu, Z. Yi, X. Lin, and P. Abbeel,
``GELLO: A general, low-cost, and intuitive teleoperation framework for robot manipulators,''
in \emph{Proc. IEEE/RSJ Int. Conf. Intell. Robots Syst. (IROS)}, 2024, pp. 12156--12163.

\bibitem{liu2025factr}
J. J. Liu, Y. Li, K. Shaw, T. Tao, R. Salakhutdinov, and D. Pathak,
``FACTR: Force-attending curriculum training for contact-rich policy learning,''
in \emph{Proc. Robotics: Science and Systems (RSS)}, 2025.

\bibitem{tobergte2011sigma7}
A. Tobergte \emph{et al.},
``The sigma.7 haptic interface for MiroSurge: A new bi-manual surgical console,''
in \emph{Proc. IEEE/RSJ Int. Conf. Intell. Robots Syst. (IROS)}, 2011, pp. 3023--3030.

\bibitem{du2021hri}
Z. Du, Y. Liang, Z. Yan, L. Sun, and W. Chen,
``Human-robot interaction control of a haptic master manipulator used in laparoscopic minimally invasive surgical robot system,''
\emph{Mech. Mach. Theory}, vol. 156, Art. no. 104132, 2021.

\bibitem{kang2022workspace}
D. Kang and D.-S. Kwon,
``An ergonomic comfort workspace analysis of master manipulator for robotic laparoscopic surgery with motion scaled teleoperation system,''
\emph{Int. J. Med. Robot. Comput. Assist. Surg.}, vol. 18, no. 6, Art. no. e2448, 2022.

\bibitem{wong2023ergonomics}
S. W. Wong, Z. H. Ang, R. Lim, X. J. Wong, and P. Crowe,
``Factors affecting upper limb ergonomics in robotic colorectal surgery,''
\emph{J. Surg. Case Rep.}, vol. 2023, no. 11, Art. no. rjad632, 2023.

\bibitem{wong2024manipulation}
S. W. Wong and P. Crowe,
``Manipulation ergonomics and robotic surgery---a narrative review,''
\emph{Ann. Laparosc. Endosc. Surg.}, vol. 9, Art. no. 15, 2024.

\bibitem{fang2024airexo}
H. Fang \emph{et al.},
``AirExo: Low-cost exoskeletons for learning whole-arm manipulation in the wild,''
in \emph{Proc. IEEE Int. Conf. Robot. Autom. (ICRA)}, 2024, pp. 15031--15038.

\bibitem{yang2024ace}
S. Yang \emph{et al.},
``ACE: A cross-platform and visual-exoskeletons system for low-cost dexterous teleoperation,''
in \emph{Proc. Conf. Robot Learn. (CoRL)}, 2024, pp. 4895--4911.

\bibitem{marinho2019}
M. M. Marinho \emph{et al.},
``A unified framework for the teleoperation of surgical robots in constrained workspaces,''
in \emph{Proc. IEEE Int. Conf. Robot. Autom. (ICRA)}, 2019, pp. 2721--2727.

\bibitem{su2020}
H. Su \emph{et al.},
``Internet of Things (IoT)-based collaborative control of a redundant manipulator for teleoperated minimally invasive surgeries,''
in \emph{Proc. IEEE Int. Conf. Robot. Autom. (ICRA)}, 2020, pp. 9737--9742.

\bibitem{audonnet2024}
F. P. Audonnet, J. Grizou, A. Hamilton, and G. Aragon-Camarasa,
``TELESIM: A modular and plug-and-play framework for robotic arm teleoperation using a digital twin,''
in \emph{Proc. IEEE Int. Conf. Robot. Autom. (ICRA)}, 2024, pp. 17770--17777.

\bibitem{su2020crtk}
Y.-H. Su \emph{et al.},
``Collaborative Robotics Toolkit (CRTK): Open software framework for surgical robotics research,''
in \emph{Proc. IEEE Int. Conf. Robot. Comput. (IRC)}, 2020, pp. 48--55.

\bibitem{rodriguez2026}
A. Rodriguez \emph{et al.},
``An open-source robotics research platform for autonomous laparoscopic surgery,''
\emph{arXiv preprint arXiv:2603.08490}, 2026.

\bibitem{wensing2017proprioceptive}
P. M. Wensing, A. Wang, S. Seok, D. Otten, J. Lang, and S. Kim,
``Proprioceptive actuator design in the MIT Cheetah: Impact mitigation and high-bandwidth physical interaction for dynamic legged robots,''
\emph{IEEE Trans. Robot.}, vol. 33, no. 3, pp. 509--522, 2017.

\bibitem{conti2005spanning}
F. Conti and O. Khatib,
``Spanning large workspaces using small haptic devices,''
in \emph{Proc. World Haptics Conf.}, 2005, pp. 183--188.

\bibitem{openhembodiment2026}
N. Nelson \emph{et al.},
``Open-H-Embodiment: A large-scale dataset for enabling foundation models in medical robotics,''
arXiv:2604.21017, 2026. Versius cholecystectomy subset. [Online]. Available:
\url{https://huggingface.co/datasets/nvidia/PhysicalAI-Robotics-Open-H-Embodiment}
(accessed Sep. 15, 2026).

\bibitem{hosseinabadi2022force}
A. H. Hadi Hosseinabadi and S. E. Salcudean,
``Force sensing in robot-assisted keyhole endoscopy: A systematic survey,''
\emph{Int. J. Robot. Res.}, vol. 41, no. 2, pp. 136--162, 2022.

\end{thebibliography}
\end{document}